\documentclass[journal]{IEEEtran}
\ifCLASSINFOpdf
\else
\fi
\usepackage{graphicx}
\usepackage{url}
\usepackage{dirtree}
\usepackage{amssymb}
\usepackage{amsmath}
\usepackage{hyperref}
\hypersetup{
    colorlinks=true,
    urlcolor=blue,
    pdfpagemode=FullScreen,
    }
\usepackage{amsmath}
\usepackage{amssymb}
\begin{document}
%
\title{InLUT3D: Challenging real indoor dataset for point cloud analysis}
%
%
%

\author{Jakub~Walczak
\thanks{J.Walczak is affiliated to Lodz University of Technology, Lodz, Poland.}
\thanks{Manuscript received April 19, 2005; revised September 17, 2014.}}

\maketitle

\begin{abstract}
In this paper, we introduce the InLUT3D point cloud dataset, a comprehensive resource designed to advance the field of scene understanding in indoor environments. The dataset covers diverse spaces within the W7 faculty buildings of Lodz University of Technology, characterised by high-resolution laser-based point clouds and manual labelling. Alongside the dataset, we propose metrics and benchmarking guidelines essential for ensuring trustworthy and reproducible results in algorithm evaluation. We anticipate that the introduction of the InLUT3D dataset and its associated benchmarks will catalyse future advancements in 3D scene understanding, facilitating methodological rigour and inspiring new approaches in the field.
\end{abstract}


%
\IEEEpeerreviewmaketitle

\section{Introduction}
\begin{figure}[hb]
    \centering
    \includegraphics[width=0.5\textwidth]{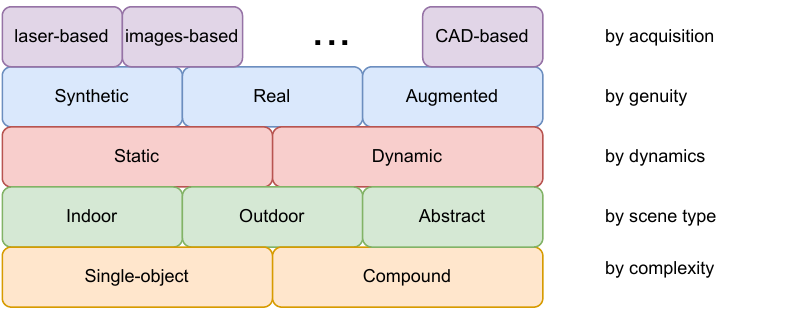}
    \caption{A sample taxonomy of point clouds}
    \label{fig:pt_tax}
\end{figure}

%
%
%
%
\IEEEPARstart{D}{ata} are an indispensable cornerstone of machine learning methods. They are particularly relevant in the domain of artificial neural networks where a generative model to be revealed is implicit and concealed in the network's architecture. Among all forms of data, volumetric entities --- point clouds --- are garnering increasing attention due to the more comprehensive structural information they might hold in comparison to other formats. 
Point cloud datasets can be categorised in the following manner (Fig. \ref{fig:pt_tax}) and the selection of the appropriate dataset always depends on the undertaken task and expected results. 

In the domain of point clouds, the primary tasks comprise: object classification \cite{su2015multi,dgcnn,walczak2021vicinity,jakub2021cva}, object detection \cite{li2016vehicle,lang2019pointpillars,shen2023v,rukhovich2023tr3d}, and semantic or instance segmentation \cite{walczak2023ultrasmall,wu2023point,zhu2023ponderv2,wang2023octformer, schult2023mask3d}. Due to significant annotation expenses and to assure consistent bench-marking, the advances in those areas require publicly available datasets. Fortunately, the growing accessibility of data and software, also actively promoting by the European Commission \cite{ec_os}, causes growth in their number. WPC \cite{9756929}, Tinto \cite{afifi_ahmed_j_m_2023_2256}, Real3D-AD \cite{liu2023real3d}, or Building3D \cite{wang2023building3d} are just several notable point cloud datasets released recently. 

In this paper, we present a novel laser point cloud dataset titled \textit{InLUT3D} (Indoor Lodz University of Technology rooms) version 1.0. It represents a challenging indoor dataset comprising scans of diverse indoor environments from different acquisition point inside the LUT W7 faculty buildings. 

The InLUT dataset was released for non-commercial purposes (Creative Commons Attribution Non Commercial 2.5 Generic license)  in ZENODO\footnote{The InLUT dataset is available under the following address: \url{https://zenodo.org/doi/10.5281/zenodo.81314867}.} platform.

\section{Related Datasets}
In this section, we present the two most relevant datasets of static point clouds. 


\subsection{S3DIS}
One of the well-known indoor datasets is the one released by Stanford University, commonly referred to as S3DIS \cite{2017arXiv170201105A}. This dataset comprises over 270 point clouds spread across six large areas. Each point is described by its Cartesian coordinates and color. The dataset includes labels for 13 unequally distributed categories (Fig. \ref{fig
}). The dataset was constructed based on reconstructions from RGB-D images acquired using a Matterport Camera.

\begin{figure}
    \centering
    \includegraphics[width=0.4\textwidth]{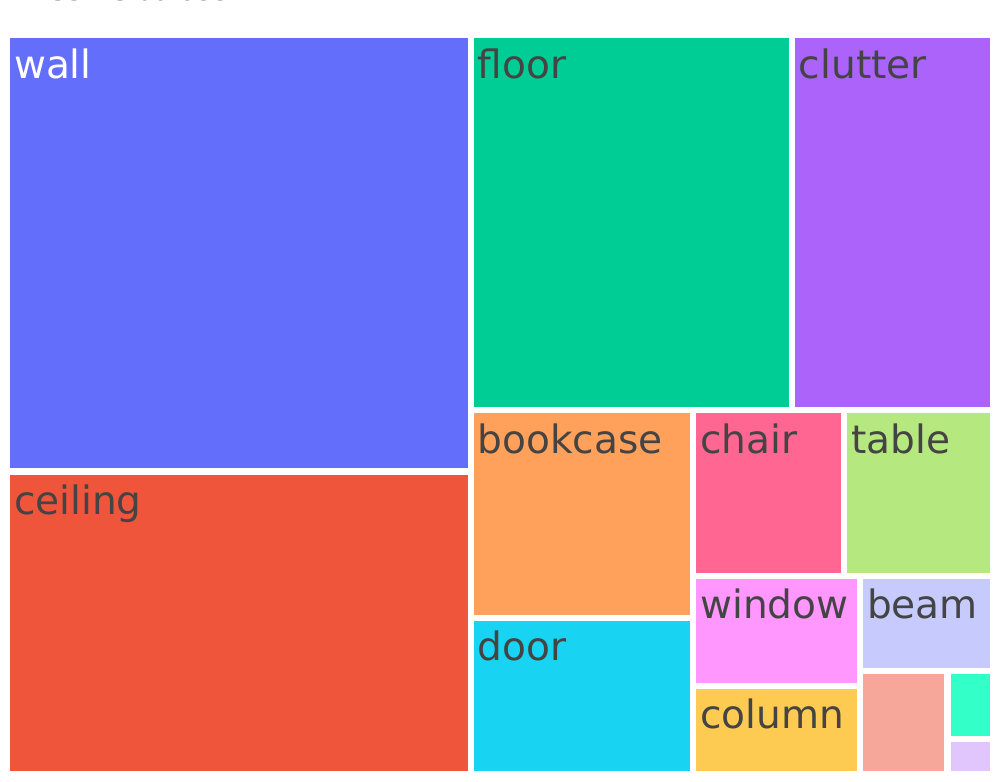}
    \caption{Distribution of points' categories in S3DIS dataset.}
    \label{fig:s3dis_dist}
\end{figure}

\subsection{Semantics3D}
Semantics3D \cite{hackel2017isprs}, on the other hand, is an outdoor dataset. It contains only 15 training samples and an additional 15 for testing purposes. Despite the limited number of point clouds, each contains a large number of points, as they were acquired using a precise scanning device.

\begin{figure}
\centering
\includegraphics[width=0.4\textwidth]{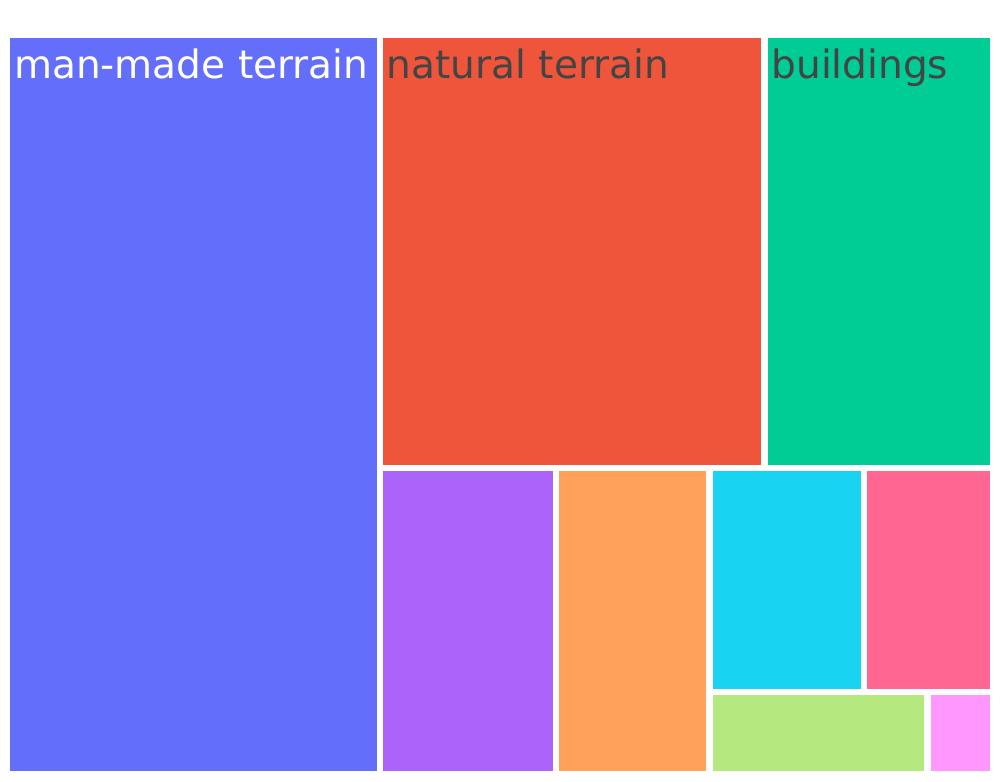}
\caption{Distribution of point categories in the Semantics3D dataset.}
\label{fig
}
\end{figure}

The points are distributed among eight categories (Fig. \ref{fig
}). Each point contains Cartesian coordinates, intensity, and colour.





\section{InLUT dataset}
The InLUT dataset is a static indoor point cloud dataset comprising 321 scenes acquired across three buildings at the Faculty of Technical Physics, Information Technology, and Applied Mathematics of Lodz University of Technologies. The dataset covers various indoor environments such as lecture halls, lecturer's offices, corridors, and toilets. In total, the InLUT dataset contains  $>3.5 \cdot 10^9$ points, each described by its Cartesian coordinates, un-normalised RGB colour values (in the range [0, 255]), category code, and instance identifier (Eq. \ref{eq
}).

\begin{equation}
p_j \in \mathcal{P} \land p_j = \{x, y, z, R, G, B, \mathcal{C}, \mathcal{I}\},
\label{eq
}
\end{equation}
where $x$,$y$,$z$ represent the Cartesian coordinates of a point relative to the origin of the coordinate system located at the centre of the laser head, $\mathcal{C}$ is a category code associated with the point (e.g., ceiling), and $\mathcal{I}$ is an instance identifier indicating which object the point belongs to.

\subsection{Data Acquisition \& Preprocessing}

Point clouds were collected with the Leica BLK360 1st class laser scanner. It features a wide field of view ($360^\circ$ horizontally and $270^\circ$ vertically) and 3D point accuracy reaching barely 6 mm at 10 m (8 mm @ 20 m) \cite{blk360}. Points are coloured based on a high-resolution 150 Mpx panoramic image of the scene. The device allows the user to select one out of three levels of granularity for the resulting point cloud. For the InLUT3D dataset, we selected the level corresponding to approximately 10,000,000 points per scan.

\begin{figure}[ht]
    \centering
    \includegraphics[width=0.4\textwidth]{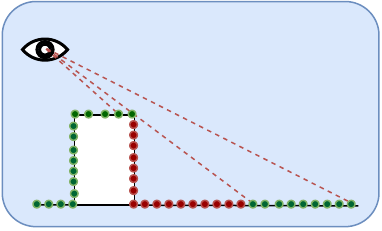}
    \caption{Visibility of points for single-setup point clouds. Green points are visible to the laser beam, red points are eclipsed.}
    \label{fig:scan}
\end{figure}

More complex indoor environments are scanned from several locations (two to four) to cover as much of the scene as possible. Trivial rooms are fully contained in single setup.
Scans are not merged for a single room. Each scan is treated as a separate dataset sample (single-setup point cloud). A single sample contains a structured point cloud with no overlaps but with missing scene elements --- "scanning shadows" (Figure \ref{fig:scan}).

Collected data were then preprocessed with the Leica Cyclone REGISTER software. The preprocessing stage relied on converting the BLK360-specific binary format to the PTS textual format. We chose the textual format to facilitate data preview. We opted for PTS format due to its popularity and ease of manipulation. Furthermore, we tested several binary formats to evaluate possible space reduction, but the results showed almost no reduction and in some cases, the database size increased (likely due to specific data organization or extra metadata).

\subsection{Data Labelling \& Postprocessing}

\begin{figure}[ht]
    \centering
    \includegraphics[width=0.49\textwidth]{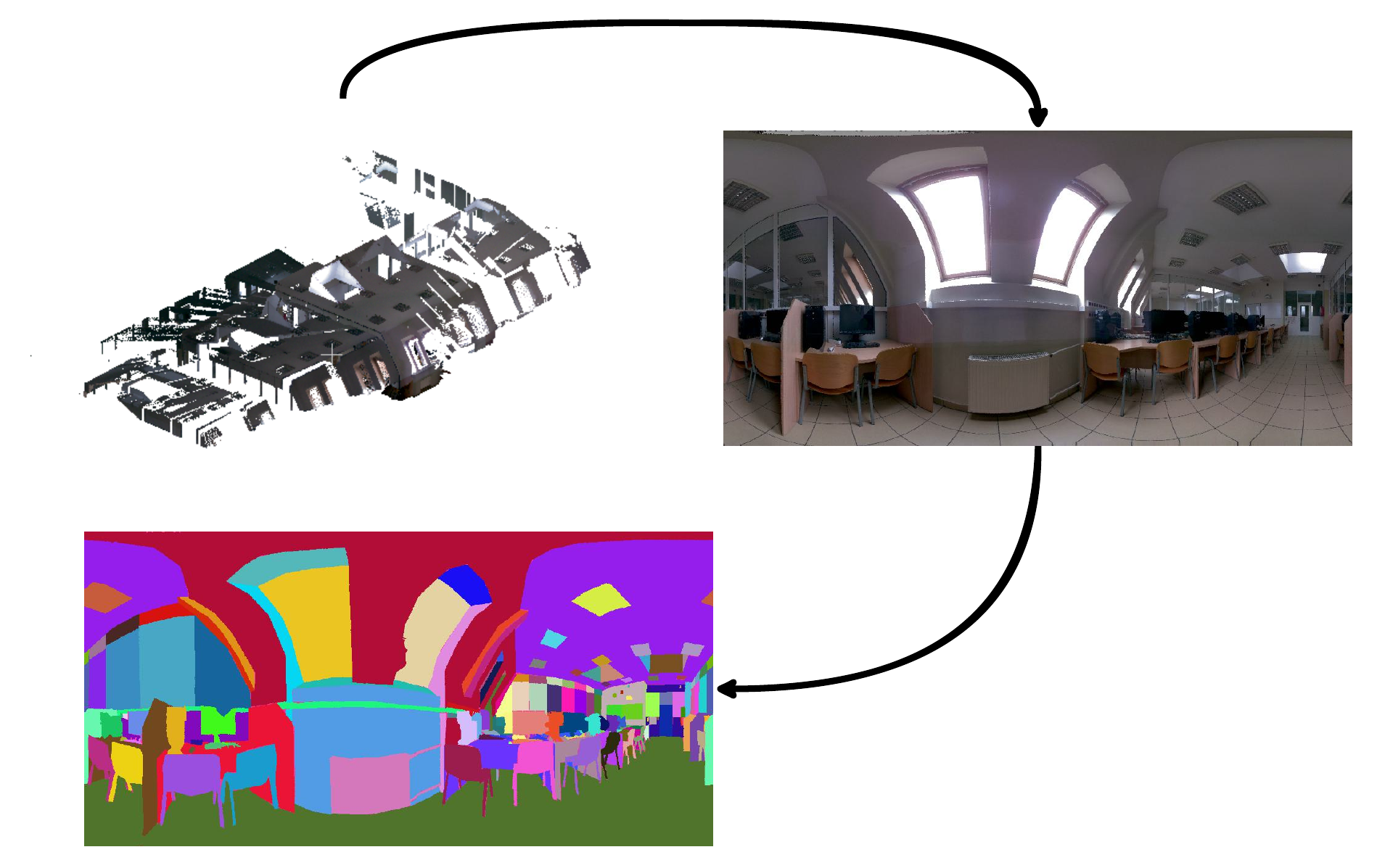}
    \caption{Point cloud converted to spherical projection.}
    \label{fig:projection}
\end{figure}

Each point cloud in the InLUT dataset was manually labeled using spherical projection of high resolution (Figure \ref{fig:projection}). Points belonging to a specific instance of a category were marked by selecting corresponding pixels.

Points were grouped into 18 main categories (Table \ref{tab:cats}) and an additional `unknown` label for points that were difficult to classify.

\begin{table}[ht]
\centering
\caption{Categories of points in the InLUT dataset.}
\label{tab:cats}
\begin{tabular}{|l|l|l|l|l|l|}
\hline
ceiling & floor & wall & stairs & column & chair \\ \hline
sofa & table & storage & door & window & plant \\ \hline
dish & wall-mounted & device & radiator & lighting & other \\ \hline
\end{tabular}
\end{table}
\subsection{Object Class Statistics}

\begin{figure}[ht]
    \centering
    \includegraphics[width=0.49\textwidth]{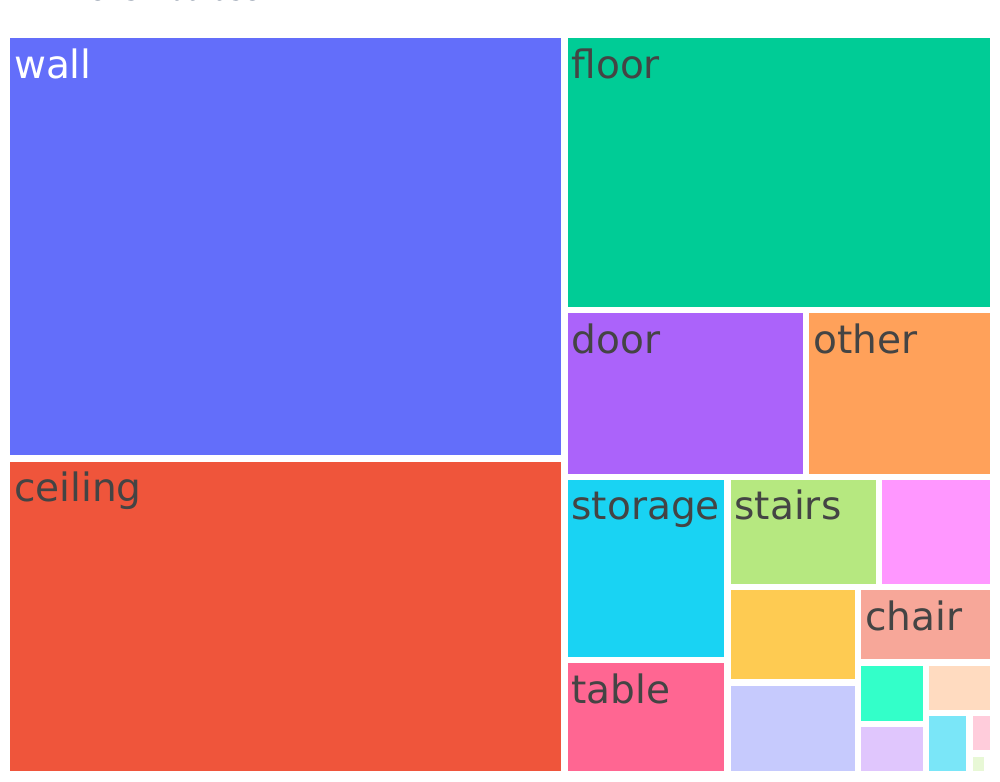}
    \caption{Distribution of points' categories in the InLUT3D dataset.}
    \label{fig:inlut}
\end{figure}

In this chapter, we present statistics related to objects included in the InLUT3D dataset. The contribution of each class in the dataset is depicted in the tree map in Fig. \ref{fig:inlut}.

Similar to the S3DIS dataset, the majority of points are labeled as walls, floors, or ceilings. This is expected as these surfaces are common and densely covered by points. Each category has a significant number of instances. The number of instances for each category is presented in Table \ref{tab:inlut3d}.

\begin{table}[ht]
\centering
\caption{Number of instances by category in the InLUT3D dataset.}
\label{tab:inlut3d}
\begin{tabular}{|l|l|}
\hline
\textbf{Category} & \textbf{Number of Instances} \\ \hline \hline
ceiling & 789 \\ \hline
floor & 121 \\ \hline
wall & 3,562 \\ \hline
door & 516 \\ \hline
wall mounted & 2,536 \\ \hline
lighting & 1,815 \\ \hline
column & 182 \\ \hline
chair & 2,610 \\ \hline
storage & 594 \\ \hline
window & 1,437 \\ \hline
device & 1,109 \\ \hline
radiator & 26 \\ \hline
table & 845 \\ \hline
plant & 121 \\ \hline
dish & 60 \\ \hline
stairs & 8 \\ \hline
other & 5,426 \\ \hline
not classified & 315 \\ \hline
\end{tabular}
\end{table}

\subsection{Dataset Organisation}

In order to facilitate trustworthy and reliable comparisons between methods, the InLUT3D dataset is divided into two predefined splits: the \textbf{train split} and the \textbf{test split}. Researchers may choose to use a validation split extracted from the training data as needed.

The InLUT3D dataset is organised in an intuitive manner: each point cloud is contained in a separate directory named following the convention \textit{setup\_nbr}, where \textit{nbr} is a successive number of a setup. Within each setups' directory, alongside the point cloud in PTS format, there are two JPG images: one showing the spherical projection of the room in its original RGB colours and another showing the segmented view. Additionally, the dataset directory includes a \textit{labels.json} file defining all the categories with their numerical codes and associated unique RGB colours. Two extra text files, \textit{train.txt} and \textit{test.txt}, list the directories included in the predefined \textbf{train} and \textbf{test} splits respectively. A simple directory structure tree is presented below:

\dirtree{%
.1 /inlut3d.zip.
.2 labels.json.
.2 setup\_0/.
.3 setup\_0.pts.
.3 segmentation.jpg.
.3 projection.jpg.
.2 setup\_1/.
.3 setup\_1.pts.
.3 segmentation.jpg.
.3 projection.jpg.
.2 \ldots.
}

\section{Benchmarking}

For reliable benchmarking, we recommend researchers use well-established measures averaged across all samples in a split.

Assuming the subset of InLUT3D is denoted as $\mathbb{D}$, a single point cloud as $\mathbf{s}$, a set of categories as $\mathbf{L}$, with true labels as $y$ and predicted labels as $\hat{y}$, the recommended measures are presented below.

For object classification: \textbf{Overall Accuracy} (Eq. \ref{eq:oa}) and \textbf{Mean Accuracy} (Eq. \ref{eq:ma}).

\begin{equation}
OA(\mathbf{s}) = \frac{1}{|\mathbf{s}|} \sum_{i=1}^{|\mathbf{s}|} \mathbf{1}(y_i = \hat{y}_i),
\label{eq:oa}
\end{equation}

\begin{equation}
mA(\mathbf{s}) = \sum_{k \in \mathbf{L}} \frac{\sum_{i=1}^{|\mathbf{s}|} \mathbf{1}(y_i = \hat{y}_i \land y_i = k)}{\sum_{i=1}^{|\mathbf{s}|} \mathbf{1}(y_i = k)},
\label{eq:ma}
\end{equation}

where $\mathbf{1}(\cdot)$ is the indicator function.

For semantic and instance segmentation: \textbf{Jaccard Score} (Eq. \ref{eq:jaccard}) and \textbf{Mean Average Precision} (mAP). Mean Average Precision is defined as in \cite{hirling2024segmentation} --- the average of Average Precision (AP) scores across all categories.

\begin{equation}
J(\mathbf{s}) = \frac{|y \cap \hat{y}|}{|y \cup \hat{y}|},
\label{eq:jaccard}
\end{equation}

The AP score for a single category $k$ is given by:

\begin{equation}
AP(k) = \int p_k(r_k) dr,
\label{eq:mAP1}
\end{equation}

and mean Average Precision is defined as:

\begin{equation}
mAP = \sum_{k \in \mathbf{L}} AP(k).
\label{eq:mAP2}
\end{equation}

The resulting metrics for the dataset are presented as the arithmetic mean of corresponding metrics obtained for each sample in the dataset (Eq. \ref{eq:res}).

\begin{equation}
\mathcal{M} = \frac{1}{|\mathbb{D}|} \sum_{\mathbf{s} \in \mathbb{D}} m(\mathbf{s}),
\label{eq:res}
\end{equation}

where $\mathcal{M}$ is the resulting metric value for the dataset and $m$ is one of the listed metrics (i.e., $OA$, $mA$, $J$, or $mAP$).

The above metrics can be easily computed using \textit{torchmetrics} \cite{detlefsen2022torchmetrics} library. The following setups should be applied:

\begin{enumerate}
    \item To compute \textbf{overall accuracy} the class \texttt{MulticlassAccuracy} with arguments:
    \begin{itemize}
        \item \texttt{num\_classes} set to \textbf{18},    
        \item \texttt{average} set to \textbf{micro},
        \item \texttt{multidim\_average} set to \textbf{global}
    \end{itemize}

    \item To compute \textbf{mean accuracy} the class \texttt{MulticlassAccuracy} with arguments:
    \begin{itemize}
        \item \texttt{num\_classes} set to \textbf{18},    
        \item \texttt{average} set to \textbf{macro},
        \item \texttt{multidim\_average} set to \textbf{global}
    \end{itemize}    

    \item To compute \textbf{Jaccard index} the class class \texttt{MulticlassJaccardIndex} with arguments:
    \begin{itemize}
        \item \texttt{num\_classes} set to \textbf{18},    
        \item \texttt{average} set to \textbf{macro},
    \end{itemize}    

    \item To compute \textbf{average precision} score the class \texttt{MulticlassAveragePrecision} with arguments:
    \begin{itemize}
        \item \texttt{num\_classes} set to \textbf{18},    
        \item \texttt{average} set to \textbf{macro},
    \end{itemize}     
\end{enumerate}







\section{Conclusion}
In this paper, we introduced the InLUT3D dataset, a comprehensive collection of static indoor point clouds obtained using the Leica BLK360 laser scanner within the W7 faculty buildings of Lodz University of Technology. This dataset captures diverse indoor environments, including lecture halls, offices, corridors, and other spaces, providing a rich resource for research in indoor scene understanding.

We outlined the dataset's structure and provided guidelines for robust method validation, emphasising the importance of predefined train and test splits in results reproducibility. These measures ensure the reliability of results when evaluating algorithms on the InLUT3D dataset.

In summary, the InLUT3D dataset serves not only as a valuable benchmark for assessing indoor scene understanding algorithms but also contributes to advancing research in 3D perception and computer vision. We anticipate that this dataset and its associated benchmarking framework will stimulate new approaches and innovations in the field, fostering further advancements in indoor 3D scene analysis.


%


\section*{Acknowledgment}
This work has received fundings from the Polish National Centre for Research and Development under the LIDER XI program [grant number 0092/L-11/2019, "Semantic analysis of 3D point clouds"].

The calculations mentioned in this paper ware performed using the BlueOcean computational cluster which is part of TUL Computing \& Information Services Center infrastructure.

\ifCLASSOPTIONcaptionsoff
  \newpage
\fi



\bibliographystyle{IEEEtran}
\bibliography{content}
%

%


\begin{IEEEbiography}[{\includegraphics[width=1in,height=1.25in,clip,keepaspectratio]{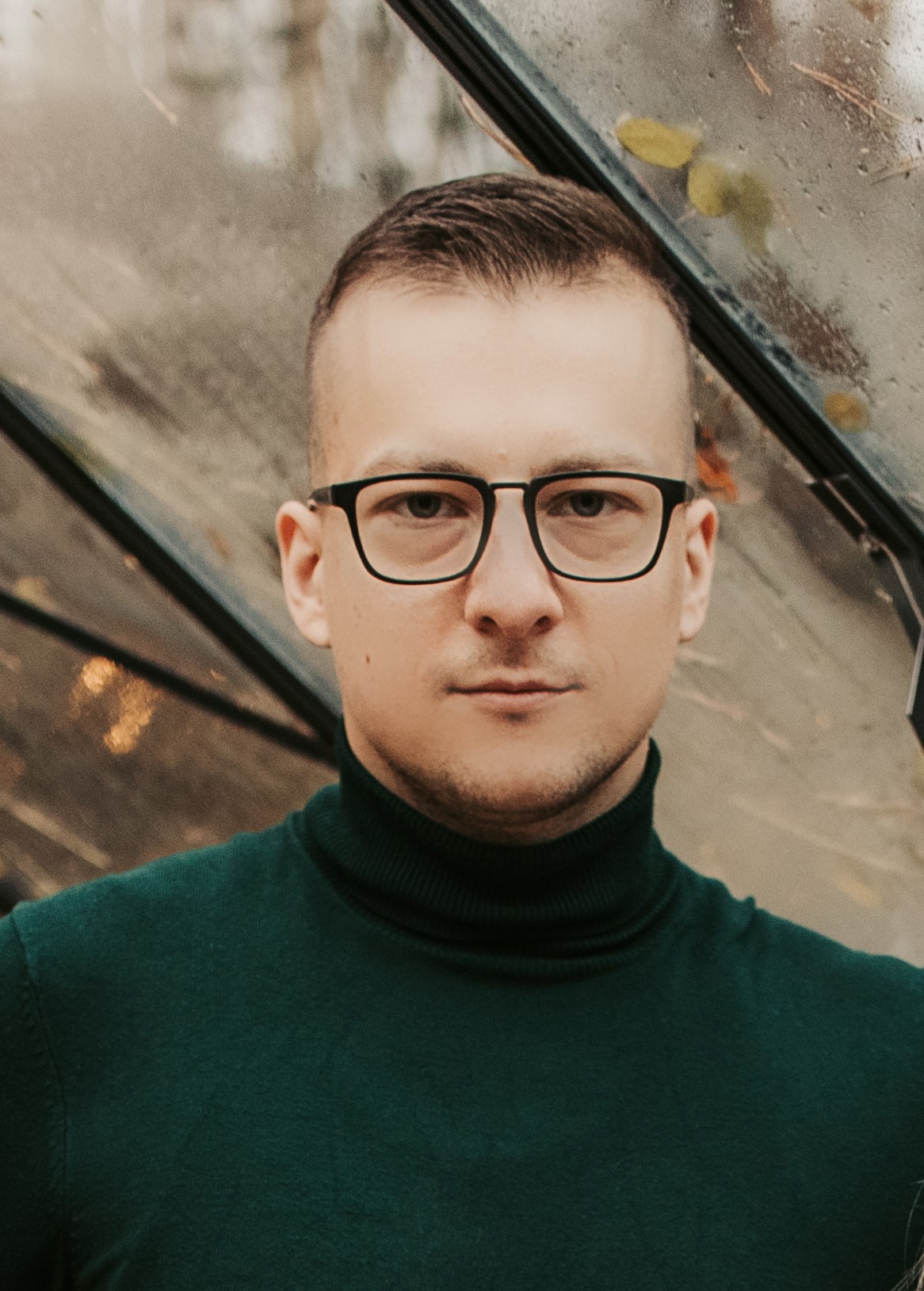}}]{Jakub Walczak}
defended his Ph.D. in 2022 on Lodz University of Technology. Since 2017, he an assistant teacher therein. He worked as a consultant of scientific software development in CMCC Foundation. Since January 2024 he cooperates with OpenNebula Systems in the area of AI-based solutions. He is an author of publications in the area of machine learning and point cloud processing as well as he delivered several presentations during international conferences. His research interests include artificial intelligence, computer vision, and big data management. Currently, he manages the research project "Semantic analysis of 3D point clouds" aiming at the enhancement of state-of-the-art methods for point cloud processing. 
\end{IEEEbiography}



\vfill




\end{document}